\documentclass[sn-mathphys-num]{sn-jnl}

\usepackage{graphicx}%
\usepackage{multirow}%
\usepackage{amsmath,amssymb,amsfonts}%
\usepackage{amsthm}%
\usepackage{mathrsfs}%
\usepackage[title]{appendix}%
\usepackage{xcolor}%
\usepackage{textcomp}%
\usepackage{manyfoot}%
\usepackage{booktabs}%
\usepackage{listings}%
\usepackage{subcaption}
\usepackage{booktabs, tabularx}
\usepackage{lineno,hyperref}
\usepackage{subcaption}
\usepackage{multirow}
\usepackage{footnote}
\usepackage{longtable}
\usepackage{array}  
\usepackage{comment}
\usepackage{makecell}
\usepackage{diagbox}
\usepackage{pifont}
\usepackage{rotating}
\usepackage{pifont}
\usepackage{float}

\usepackage{natbib}

\newcommand{\xmark}{\ding{55}}%

\theoremstyle{thmstyleone}%
\theoremstyle{thmstyletwo}%

\theoremstyle{thmstylethree}%

\begin{document}

\title[Article Title]{Understanding User Preferences in Explainable Artificial Intelligence: A Survey and a Mapping Function Proposal.}


\author*[1]{\fnm{Maryam} \sur{Hashemi}}\email{m.hashemi@unsw.edu.au}

\author[1]{\fnm{Ali} \sur{Darejeh}}\email{ali.darejeh@unsw.edu.au}

\author[1, 2]{\fnm{Francisco} \sur{Cruz}}\email{f.cruz@unsw.edu.au}

\affil*[1]{\orgdiv{School of Computer Science and Engineering}, \orgname{University of New South Wales}, \orgaddress{\street{Kensington}, \city{Sydney}, \postcode{2033}, \state{NSW}, \country{Australia}}}

\affil[2]{\orgdiv{Escuela de Ingeniería}, \orgname{Universidad Central de Chile}, \orgaddress{ \state{Santiago}, \country{Chile}}}



\abstract{The increasing complexity of AI systems has led to the growth of the field of Explainable Artificial Intelligence (XAI), which aims to provide explanations and justifications for the outputs of AI algorithms. 
While there is considerable demand for XAI, there remains a scarcity of studies aimed at comprehensively understanding the practical distinctions among different methods and effectively aligning each method with users’ individual needs, and ideally, offer a mapping function which can map each user with its specific needs to a method of explainability.
This study endeavors to bridge this gap by conducting a thorough review of extant research in XAI, with a specific focus on Explainable Machine Learning (XML), and a keen eye on user needs. Our main objective is to offer a classification of XAI methods within the realm of XML, categorizing current works into three distinct domains: philosophy, theory, and practice, and providing a critical review for each category. Moreover, our study seeks to facilitate the connection between XAI users and the most suitable methods for them and tailor explanations to meet their specific needs by proposing a mapping function that take to account users and their desired properties and suggest an XAI method to them. This entails an examination of prevalent XAI approaches and an evaluation of their properties.
The primary outcome of this study is the formulation of a clear and concise strategy for selecting the optimal XAI method to achieve a given goal, all while delivering personalized explanations tailored to individual users. 
}




\maketitle

\section{Introduction} \label{intro}

Artificial Intelligence (AI) algorithms have great potential for use in our daily lives. Due to the complexity and black-box nature of these algorithms, there have been numerous studies in the field of Explainable AI (XAI). The motivation behind XAI is to elucidate how an AI algorithm works in a human-understandable way. Although XAI is highly in demand, there are limited studies on understanding the differences between various methods in practice and mapping each specific method to users based on their needs~\cite{brennen2020people},~\cite{langer2021we}.

This work is an observational study on users of XAI, desired properties for XAI methods, and the missions that XAI pursues. Through this paper, we aim to answer the following questions: Who are the users of XAI, what do they want from XAI, and how should we deliver it? To achieve this, we exploring the roots and milestones of this field. This exploration will aid us in identifying shortcomings more effectively and conducting a review of recent papers in the field of XML (also called XAI in this paper). This review primarily centers on analyzing the practical application of mathematical approaches in XAI and their inherent characteristics. It explores how these attributes can be leveraged by users of XAI to effectively address their specific needs.

To explain the AI black box, two general strategies have been identified. The first approach involves replacing the black-box algorithm with a data-driven white-box algorithm such as linear regression and decision trees. In some contexts, white-box algorithms, being simpler than black-box ones, exhibit lower performance~\cite{loyola2019black},~\cite{linardatos2020explainable}. The second approach entails opening the black-box algorithm and justifying the system's output. This work is dedicated to revising the second approach.

We have categorized research in XAI into three main areas: philosophy, theory, and practice. In the philosophy category, efforts are made to define an explanation and understand how the human mind processes and comprehends explanations. The theory section focuses on providing the mathematical underpinnings of explanations and formulating the problem. The practice category delves into the practical challenges of implementing XAI.

The next section will review the previous works from the philosophy, theory, and practice perspectives. 
 Section~\ref{map} is dedicated to explaining our proposed mapping function and is our main contribution. In Section~\ref{short}, we discuss the shortcomings and possible future directions in XAI. In the last section we discuss the conclusions of our work.

\section{Previous Works}
\label{pre}
In this section, we delve into related works in the realm of XAI through three distinct perspectives. First, we explore the philosophical aspect, explaining how researchers have defined explanations within the realm of AI and discerning the characteristics of different types of explanations. Following this, we discuss the theoretical dimension, where we highlight important mathematical algorithms designed to elucidate various machine learning algorithms. Finally, we examine works that assess XAI in practical applications, define desired properties for it, and underscore critical concerns.

\subsection{Philosophy Perspective}

In this category of studies, scholars have delved into the concept of explanations and attempted to establish definitions for terms such as ``explanations," ``transparency," and ``counterfactual explanation" from the computer science perspective. In general, studies contribute to answering the following fundamental questions in this area: 1) What is an explanation? 2) What is explainable AI? 3) What are counterfactual explanations? We answer the mentioned questions in the following:

\subsubsection{ What is an explanation?}

The Oxford English dictionary defines an explanation as a statement that makes something clear and gives a reason or justification for an action. In the literature, Von Wright~\cite{Wright1971-WRIEAU} compares explanations with predictions. Predictions gaze into the future from our present state, projecting what will happen. On the contrary, explanations trace the path from the current state of affairs to events of the past. Xu et al.~\cite{xu2019explainable} provided insights into the basic question of what constitutes a valid explanation from a scientific perspective. They identified two components for a valid explanation: the object that we try to explain and the content related to the object that explains. On the other hand, O'Hara~\cite{o2020explainable} elucidates that explanations are more near to relatives than mere sentences with specific components. They discuss the idea that explanations are inherently relative to their audiences, tailored to ensure comprehension by those who receive them. O'Hara argues that explanations should be viewed as a dynamic process aimed at fostering understanding and grasping knowledge about phenomena, rather than mere textual constructs or collections of components.  By evaluating the mentioned works in this section we can summarize, the definition of ``explanation" can vary significantly depending on factors such as its intended purpose, the audience, and the expected depth of reasoning, regardless of these variables, explanations can be conceptualized as a series of established facts from the past that culminate in a reasoned conclusion about a particular outcome.

\subsubsection{What is explainable AI?}

Explainability in AI is indeed not a novel concept; its roots can be traced back to the developing stages of AI, particularly during the utilization of abductive reasoning in expert AI systems~\cite{goebel2018explainable}. It emerged as a fundamental aspect of AI design, serving as a reasoning architecture to support and elucidate the intricate workings of AI systems. Thus, explainability in AI can be considered as old as early AI itself, evolving as a natural consequence of the need to comprehend and justify the decisions made by AI systems. Hoffman et al.~\cite{https://doi.org/10.48550/arxiv.1812.04608} contend that the foundation of explainable AI stems from a fundamental necessity. Just as decision-makers can grasp and justify a computational system built upon a straightforward statistical model, there is an analogous requirement for machine learning systems. It is imperative to elucidate Machine Learning (ML) models to decision-makers, enabling them to understand and endorse the system's rationale and fairness, ultimately securing their approval.

While certain contemporary ML algorithms generate rules regarding data features to inform classification decisions, it is crucial to recognize that these rules primarily illuminate the mechanics of decision-making rather than the underlying rationale behind them. Moreover, additional artifacts of learning systems, such as annotations and visualizations, often necessitate human-driven post-processing within their distinct lines of reasoning. Thus, while these tools contribute to understanding the ``how" of decision-making, they may not necessarily provide insights into the ``why" behind those decisions without additional human interpretation and analysis; so, truly explainable systems are those that having automated reasoning as an integral part of the output without requiring human post-processing~\cite{doran2017does}.
Doran et al.~\cite{doran2017does} propose three distinct perspectives from which XAI can be approached:
 a) Opaque Systems: in these systems, the mechanisms responsible for mapping inputs to outputs are concealed from the user, rendering them invisible and inscrutable, b) Interpretable Systems: unlike opaque systems, interpretable systems allow users not only to observe but also to study and comprehend the mathematical mappings from inputs to outputs, facilitating a deeper understanding of the system's inner workings, and c) Comprehensible Systems: these systems go beyond mere interpretation by emitting symbols along with their outputs. These symbols enable users to establish connections between the properties of the inputs and their corresponding outputs, enhancing the overall comprehensibility of the system. Xu et al.~\cite{xu2019explainable} introduced a general principle suggesting an inverse relationship between the explainability of machine learning models and their accuracy. This implies that as the accuracy of a model improves, its explainability tends to decrease. However, the authors did not delve into scenarios where this principle may not hold true such as non data-driven white-box models (e.g. phenomenological models).

Hoffman et al.~\cite{https://doi.org/10.48550/arxiv.1812.04608} went beyond mere explanations and endeavored to define mental models, specifically focusing on user understanding, within the realm of XAI. Their study focused on metrics aimed at ensuring users effectively comprehend explanations, or, put differently, that explanations meet a certain standard of quality. To this end, the authors assembled a set of trigger questions essential for designing an explanation algorithm, such as ``What do I do if it gets it wrong?" Additionally, they advocated for validating the content of explanations, measuring trust within the context of XAI, and establishing empirical assertions as crucial components of this process.
Sovrano et al.~\cite{sovrano2023objective} aimed to measure and formulate the level of text explainability, drawing on ordinary language philosophy. The authors introduced a metric named Degree of Explainability (DoX). This metric posits that the degree of explainability is directly proportional to the number of relevant questions that a piece of information can correctly answer, encompassing questions such as how, when, why, and what if.

In summary, XAI endeavors to provide methodologies that not only ``justify" and ``reason", but also ``explain" about the outcomes produced by AI systems. By prioritizing clarity and comprehensibility, XAI aims to enhance trust, facilitate human understanding, and foster collaboration between AI systems and their human counterparts~\cite{goebel2018explainable}-\cite{sovrano2023objective}.

\subsubsection{What are counterfactual explanations?}

 Counterfactual (CF) explanations represent a distinct subset within the realm of XAI facilitating a connection between potential alternative outcomes resulting from alterations to the model's input~\cite{verma2020counterfactual}. Counterfactual Explanations (CFEs) involve perturbing features to demonstrate the potential outcomes under those conditions. In essence, they offer ``what-if" explanations for model outputs~\cite{mothilal2020explaining}. Based on this, we can say CF examples can be seen as more human-interpretable explanations, similar to how children learn through counterfactual examples~\cite{wachter2017counterfactual},~\cite{weisberg2013pretense},~\cite{buchsbaum2012power}. Byrne~\cite{byrne2019counterfactuals} discussed the counterfactual structure, relations, content, and inferences and mentioned that counterfactuals can be additive or subtractive. In additive counterfactuals, ``a new situation that could have happened but didn't" will be added to explanations. For example, in the case of self-driving cars, we can say, ``if the car had detected the pedestrian earlier and braked, the passenger would not have been injured.” On the other hand, subtractive counterfactuals delete a fact that already happened to reach the desired output, ``if the car had not swerved and hit the wall, the passenger would not have been injured.” The authors made the argument to conclude that we have various types of explanations with different impacts. Still, to maximize the effect of CFs, computer science needs to seek help from cognitive science.
 

Counterfactuals and causal inferences share a close relationship, often viewed as ``two sides of the same coin" ~\cite{Hume1896}. Counterfactuals enhance the causal connection between an action and its consequence. As for their content, according to Hume~\cite{Hume1896}, counterfactuals typically involve modifications to actions rather than inactions. Consider two clients: one has shares in company A and wants to move to another company, let's call it B. But they decided against it. They now  realize they would have been better off by \$ 1,200 if they had switched to Company B; the other client is in Company B, considering going to Company A, and they do switch it. But later they find out would have been better off by \$ 1,200 if they had kept stock in Company B. People created counterfactuals focused on the person who acted, ``if only they hadn’t switched…”~\cite{Daniel1982}.

Keane et al.~\cite{keane2021if} emphasized that CFEs can convey more information than other XAI methods. However, significant deficiencies exist in generating CFEs. Firstly, most works did not assess their proposed algorithms on real-world systems as they reported that less than 21\% of the papers tested their systems on real-world users. Moreover, Adadi et al.~\cite{adadi2018peeking} suggesting less than 5\% of the papers tested their systems on real-world users. Secondly, while many papers claim their CFEs are plausible, this is often not supported by user tests. The authors identified inadequate coverage and comparative testing as additional shortcomings in this field.

In conclusion, CFEs illuminate the necessary adjustments required in the features and attributes of a machine learning model's input to yield a different outcome. The primary advantage of utilizing this form of explanation, compared to others, is its potential for enhanced user-friendliness and information conveyance.
 

\subsection{Theory Perspective}
\label{theory}

From the perspective of computer science, the theoretical underpinning of XAI holds immense significance. Previous endeavors within this realm revolve around problem formulation, mathematical elucidation, and the derivation of optimal solutions.
XAI methods can generally be classified into two groups: feature importance-based approaches and counterfactuals. The former entails algorithms aimed at revealing the significance or contribution of individual features~\cite{ribeiro2016should},~\cite{lundberg2017unified}, whereas the latter focuses on algorithms tailored to identifying changes that could potentially alter the output~\cite{wachter2017counterfactual}. We will discuss both approaches in the following:

\subsubsection{Feature Importance-based Explanations}

One notable XAI method is Local Interpretable Model-agnostic Explanations (LIME)~\cite{ribeiro2016should}. LIME falls under the category of feature importance-based methods and is designed to explain local samples. The underlying concept of LIME lies in estimating a linear classifier in the local vicinity of a given sample within a nonlinear model. This intuition can be formulated as follows:

\begin{equation}
\xi(x)=\underset{g \in G}{\operatorname{argmin}} \quad \mathcal{L}\left(f, g, \pi_{x}\right)+\Omega(g),
\end{equation}
where $\mathcal{L}\left(f, g, \pi_{x}\right)$ is a measure of how unfaithful $g$ is in approximating $f$ in the locality defined by $\pi_{x}$ and $\Omega(g)$ is the complexity of model. $\mathcal{L}\left(f, g, \pi_{x}\right)$ can be defined as follows:

\begin{equation}
\mathcal{L}\left(f, g, \pi_{x}\right)=\sum_{z, z^{\prime} \in \mathcal{Z}} \pi_{x}(z)\left(f(z)-g\left(z^{\prime}\right)\right)^{2}.
\end{equation}
$g$ is the class of linear models, such that $ g(z^{'})= w_{g} . z^{'}$ where $w_{g}$s are associated weights in the linear model and $z^{\prime}$ is the input. They used the locally weighted square loss as $\mathcal{L}$, and $\pi_{x}(z)=$ $\exp \left(-D(x, z)^{2} / \sigma^{2}\right)$ is an exponential kernel and $D$ is distance function with width $\sigma$.

\begin{figure}
    \centering
    \includegraphics[width = \textwidth]{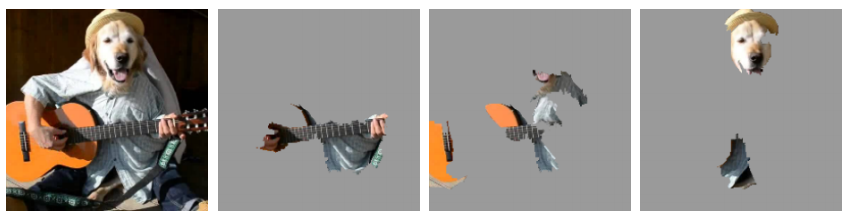}
    \caption{Explaining an image classification prediction made by Google’s Inception neural network~\cite{szegedy2015going} by LIME~\cite{ribeiro2016should} The top 3 classes predicted are “Electric Guitar” (the second picture from left), “Acoustic guitar” (the third picture from left) and “Labrador” (the last picture from left)}
    \label{lime}
\end{figure}

In Figure~\ref{lime} we display an example of LIME explanation that highlighted the super-pixels with positive weight towards a specific class, as they advocate why the model made that decision.

In the realm of feature importance approaches, Lundberg et al.~\cite{lundberg2017unified} introduced SHAP (SHapley Additive exPlanations). This approach draws inspiration from game theory, conceptualizing each feature as an agent, and aims to distribute the output fairly among these agents. In this context, a feature's importance is determined by its proportional contribution to the overall output, akin to allocating shares in a fair manner. Similar to LIME, SHAP furnishes local explanations, elucidating the contribution of each feature within a specific context. The explanation provided by SHAP is defined as follows:

\begin{equation}
  g\left(z^{\prime}\right)=\phi_{0}+\sum_{i=1}^{M} \phi_{i} z_{i}^{\prime}, 
\end{equation}
where $g$ is the explanation model, $z^{\prime} \in\{0,1\}^{M}$ is the coalition vector for agents, $M$ is the maximum coalition size, and $\phi_{i} \in \mathbb{R}$ is feature $i$ contribution. $\phi_{i}$ can be calculated as follow:

\begin{equation}
  \phi_{i}=\sum_{S \subseteq F \backslash\{i\}} \frac{|S| !(|F|-|S|-1) !}{|F| !}\left[f_{S \cup\{i\}}\left(x_{S \cup\{i\}}\right)-f_{S}\left(x_{S}\right)\right].  
\end{equation}

$F$ is the set of all features and $S$ is a subset of features ($S \in F$). $S \subseteq F \backslash\{i\}$ indicates that $S$ is any subset of $F$ while feature $i$ is not included. $f_{S \cup\{i\}}$ is the model trained with the feature present, and $f_{S}$ is trained with the feature withheld. Models are compared on the current input $f_{S \cup\{i\}}\left(x_{S \cup\{i\}}\right)-f_{S}\left(x_{S}\right)$, where $x_{S}$ represents the values of the input features in the set $S$. The main concern regarding SHAP is that this approach consider features independent and doesn't take into account casual inferences.

Two additional approaches that have garnered significant attention are Partial Dependence Plot (PDP)~\cite{friedman2001greedy} and Accumulated Local Effects (ALE)~\cite{apley2020visualizing}. Unlike previous methods, ALE is global in nature, providing insights into the effects of features on predictions in a general sense, rather than being sample-specific. The underlying ideas of PDP and ALE are explained below:

Consider we have a supervised learning model for approximating $\mathbb{E}[Y |X = x] \approx f (x)$, where $Y$ is a scalar response variable, $X = (X_{1}, X_{2}, \dots , X_{d})$ is a vector of $d$ predictors, and $f(\cdot)$ is the fitted model that predicts $Y$ (or the probability that $Y$ falls into a particular class, in the classification setting).  The training data to which the model is fit consists of $n (d + 1)$-variate observations of $\{y_{i}, x_{i} = (x_{i,1}, x_{i,2}, . . . , x_{i,d}) :i = 1, 2, \dots , n\}$. To calculate the effect of one predictor (say $X_{1}$) on the predicted response, a PD plot can be formulated as follow:
\begin{equation}
f_{1, P D}\left(x_{1}\right) \equiv \mathbb{E}\left[f\left(x_{1}, X_{2}\right)\right]=\int p_{2}\left(x_{2}\right) f\left(x_{1}, x_{2}\right) d x_{2},
\end{equation}
where $p_{2}(\cdot)$ denotes the marginal distribution of $X_{2}$.


ALE is calculated as follow:

\begin{equation}
\begin{aligned}
f_{1, A L E}\left(x_{1}\right) & \equiv \int_{x_{\min , 1}}^{x_{1}} \mathbb{E}\left[f^{1}\left(X_{1}, X_{2}\right) \mid X_{1}=z_{1}\right] d z_{1}-\text { constant } \\
&=\int_{x_{\min , 1}}^{x_{1}} \int p_{2 \mid 1}\left(x_{2} \mid z_{1}\right) f^{1}\left(z_{1}, x_{2}\right) d x_{2} d z_{1}-\text { constant },
\end{aligned}
\label{ALEEQ}
\end{equation}
where $f^{1}\left(x_{1}, x_{2}\right) \equiv \frac{\partial f\left(x_{1}, x_{2}\right)}{\partial x_{1}}$ represents the local effect of $x_{1}$ on $f(\cdot)$ at $\left(x_{1}, x_{2}\right)$, and $x_{\min , 1}$ is some value chosen near the lower bound of the effective support of $p_{1}(\cdot)$, e.g., just below the smallest observation $\min \left\{x_{i, 1}: i=1,2, \ldots, n\right\}$. Choice of $x_{\min , 1}$ is not important, as it only affects the vertical translation of the ALE plot of $f_{1, A L E}\left(x_{1}\right)$ versus $x_{1}$, and the constant in Equation~\ref{ALEEQ} will be chosen to vertically center the plot.




\subsubsection{ Counterfactual Explanations}

All the methods discussed so far are feature-based, focusing on explaining the importance of individual features. In contrast, counterfactual explanations deviate from this approach and do not rely on feature importance.
One of the early contributors to the formulation of counterfactual explanations was Wachter et al.~\cite{wachter2017counterfactual} in 2017. Wachter formulated the problem in the following manner:

\begin{equation}
	arg \min_{w} l(f_{w}(x_{i}), y_{i}) + p(w),
\end{equation}

\begin{equation}
	arg \min_{x^{'}}  \max _{\lambda} \lambda (f_{w}(x^{'})- y^{'})^{2} + d(x_{i},x^{'}),
\end{equation}
where $w$ indicates the weight, $y_{i}$ is the label for data point $x_{i}$, and $p(.)$ is a regularizer over the 
weights in an AI algorithm. We wish to find a counterfactual $x^{'}$ as close to the original point $x_{i}$ as possible such that $f_{w}(x^{'})$ is our desired output $y^{'}$. Here, $d$ is a distance function that measures how far the 
counterfactual $x^{'}$ is from the original data point $x_{i}$. Watcher considers this distance as follows, which is the median absolute deviation of feature $k$ over the set of points $P$~\cite{wachter2017counterfactual}:

\begin{equation}
	d(x_{i},x^{'})= \sum _{k \in F} \dfrac{|X_{i,k} - x^{'}_{k}|}{median_{j \in P} (|X_{j,k} - median_{l \in P}(X_{l,k})|)}.
\end{equation}

In 2018, Ustun et al.~\cite{ustun2019actionable} made a notable contribution to the field of counterfactual explanations. In their research, they referred to counterfactuals as actionable recourse. The primary distinction between Ustun's work and Wachter's~\cite{wachter2017counterfactual} was Ustun's utilization of the percentile as the distance function, with a specific focus on ensuring feasibility.
Feasibility, in this context, refers to the condition that the method should not suggest changes to immutable features, such as race, ensuring practical and realistic counterfactuals. Ustun et al. formulated their problem as follows:

\begin{align}
\begin{split}
	& \min cost(a;x)\\
	& s.t. \quad f(x+a)= 1 \\
	 & a \in A(x)
  \end{split}
\end{align}

\begin{equation}
	cost(x+a;x)= \max _{j \in J_{A}} |Q_{j}(x_{j}+a_{j})- Q_{j}(x_{j})|,
\end{equation}
where $x$ is the current data point, $x+a$ is the counterfactual point, $f(x)$ is the AI function, $A(x)$ is a set of feasible actions, and $Q_{j}(\cdot)$ is the Continuous probability distributions (CDF) of $x_{j}$ in the target population.

In~\cite{lucic2019focus}, the authors approached the challenge of finding CFEs as a gradient-based optimization task. For models that lack differentiability, such as tree ensembles, they incorporated probabilistic model approximations into the optimization framework. Notably, Euclidean, Cosine, and Manhattan distances were mentioned as potential distance metrics within the cost function. Although they tested their proposed method on some datasets, this study lack of real world user evaluation.

In 2020, the field of XAI, particularly CFEs, experienced a surge in publications. Karimi et al.~\cite{karimi2020model} introduced a model-agnostic method for discovering CFEs named MACE. The strength of their work lies in its model-agnostic nature, making it applicable to various AI algorithms, including neural networks and decision trees. The authors also emphasized the consideration of the feasibility property, referred to as plausibility in their work.
The evaluation of MACE was conducted on three prominent datasets in the field: Loan approval (adult)~\cite{dua2017uci}, Credit dataset~\cite{YEH20092473}, and Pretrial bail (COMPAS) dataset~\cite{Larson2016}. Based on their findings, the proposed method not only achieves comprehensive coverage but also generates counterfactuals at more favorable distances compared to existing optimization-based approaches. Additionally, as a second strength of their work, they highlight the proposed method's ability to inform system administrators about the potential biases inherent in the model's reliance. However, it is important to note that one drawback of their proposed method is its relatively high runtime when compared to subsequently introduced methods.

Karimi et al.~\cite{karimi2020algorithmic} presented a method for generating counterfactual explanations with limited knowledge about the causality between features. They employed a gradient-based optimization approach to identify the optimal counterfactual point.
In subsequent work, Karimi et al.~\cite{karimi2021algorithmic} introduced causal reasoning into the generation of CFEs. Shifting from recourse via the nearest counterfactual to recourse through minimal interventions, the focus of this work transitioned from explanations to interventions.

In 2021, Mohammadi et al.~\cite{mohammadi2021scaling} proposed a framework based on Mixed-Integer Programming (MIP). This framework computes the nearest counterfactual explanations for the outcomes of neural networks, ensuring both an answer (provide a CFE for all individuals) and an optimal solution (return the nearest individual) compared to gradient-based approaches. One concern relating to their work is that they developed their algorithm only based on ReLu activation function in neural networks while did not evaluate other functions.
The authors in~\cite{karimi2021survey} provided an in-depth exploration of the concept of CFEs, covering aspects such as recourse definition, causality, plausibility, actionability, diversity, and sparsity. All the mentioned properties will be discussed in detail in the next section. Causality consider the casual relationship between features, plausibility makes sure the changes recommended by CFEs are plausible, and sparsity consider the trade-off between the number of features changed and the total amount of change required to obtain the CFE. They identified three essential criteria for optimal CF explanations: perfect coverage, efficient runtime, and access.

Mothilal et al.~\cite{mothilal2020explaining} proposed a techniques to find CFE named DICE in 2020. They named two desired properties for their proposed framework, feasibility and diversity. They added these two properties to their cost function to maximize feasibility and diversity. Finally, the approach has been tested on four datasets: Adult-Income dataset~\cite{dua2017uci}, LendingClub dataset~\cite{Davenport2015}, German-Credit dataset\cite{dua2017uci}, and COMPAS dataset~\cite{Larson2016}. They reported their proposed method is able to generate CFE at different distances from the original input data for various machine learning models, such as 1 layer neural networks.
Another novel approach, proposed by Wang et al.~\cite{wang2021skyline}, introduced skyline CFEs, defining the skyline of CFs as all non-dominated changes. They addressed this problem as a multi-objective optimization over actionable features, exploring a space of CFs rather than a single data point. The algorithm was validated using three datasets: UCI Adult Dataset~\cite{dua2017uci}, Give Me Some Credit (GMSC)\cite{creditfusion2011}, and HELOC Dataset~\cite{explainablemlchallenge}. Notably, the authors did not discuss the runtime of this approach. The authors mentioned that their proposed approach achieved better performance in terms of the change in numerical features (average percentile shift) \cite{pawelczyk2020learning},\cite{ustun2019actionable} for the UCI Adult and GMSC datasets, but yielded worse results for the HELOC dataset. Additionally, it achieved better sparsity results only for the GMSC dataset.

To enhance comprehension regarding which features ought to be recommended for modification by CFEs and the corresponding actions to be taken, researchers have proposed an objective function. This function evaluates pairs of actions and prioritizes them based on feature interaction~\cite{kanamori2021ordered}. The authors emphasized the existence of asymmetric interactions among features, such as causality, and argued that the total cost of an action depends on these feature interactions. Consequently, effective counterfactual methods should provide an appropriate order for changing features. To implement this idea, the authors considered an interaction matrix to discern causality between features.
In a related work, Hada et al.~\cite{hada2021exploring} directed their attention to explaining classification and regression trees among various AI algorithms. They formulated counterfactual explanations as an optimization problem, further contributing to the evolving landscape of counterfactual research.

Verma et al.~\cite{verma2021amortized} published an article that proposed a stochastic-control-based approach to generate sequential Algorithmic Recourses (ARs) or the same CF. This algorithm permits the data point to move stochastically and sequentially across intermediate states to a final state of the CF point. The authors named their approach FASTAR and mentioned the following desiderata for their algorithm. 1- Actionability (or feasibility): this property ensures that CFE only conside mutable features to change not immutable ones (such as race). 2- Sparsity: this property searches for smaller explanations since smaller explanations are more understandable to humans~\cite{miller2019explanation}. 3- Data manifold: This property ensures that the explanation respects the training data manifold~\cite{dhurandhar2019model},~\cite{kanamori2020dace}. 4- Causal constraints: this property considers the causality between features~\cite{mahajan2019preserving}. 5- Model-agnostic: this property makes sure that CFEs will be generated regardless of knowing the ML model. 6- Amortized: an amortized approach can generate ARs for several data points without optimizing separately for each of them~\cite{mahajan2019preserving}. They tested their proposed approach on three datasets, German Credit, Adult income, and Credit Default~\cite{dua2017uci}. FASTAR reached a considerably better performance run time wise for all three datasets.

We presented all important proposed methods in XAI and their contribution, dataset they used, publicity of the dataset in Table~\ref{tab1}.

In summary, previous works focused on algorithms that are specialized in explaining specific kinds of machine learning algorithms such as decision trees, convolutional neural networks~\cite{zeng2023abs}, Long Short Term Memory (LSTM)~\cite{chaoqun2023delelstm}, monotonic neural networks~\cite{el2023cardinality}, recurrent neural networks~\cite{baier2023relinet}, and heterogeneous graph neural network~\cite{li2023towards}. From a theoretical perspective, computing counterfactual explanations for an AI algorithm involves minimizing a cost function between the current point $x$ and CF point $x^{'}$, while ensuring that the output of the AI algorithm for the CF point aligns with the desired output. Various works propose different cost functions or optimization strategies to address this problem.
For feature importance-based methods, each approach introduces a unique methodology for determining the importance or contribution of features in the prediction of machine learning algorithms.
Furthermore, by observing the literature, it is apparent that the German Credit, Adult Income, and Credit Default datasets~\cite{dua2017uci} are among the most widely used datasets to evaluate and test explanation-generating algorithms in this field. These datasets serve as standard benchmarks for assessing the effectiveness and applicability of different XAI techniques.

 \subsection{Practice Perspective}
\label{practice}

In this section, we distance ourselves from the theory of XAI and delve into its real-world applications. We discuss what constitutes a ``good explanation" and a ``bad explanation" and enumerate properties that characterize a good explanation. Additionally, we explore approaches to enhance the effectiveness of applying XAI in practical scenarios.

Herm et al.~\cite{herm2021don} conducted a user study to assess the explainability and comprehensibility of XAI-transferred Artificial Neural Networks (XANN). Their conclusion highlighted that XANN exhibit superior explainability and comprehensibility compared to other explainable models, such as support vector machine and linear regression. They mentioned the main reason that users preferred XANN over other models is that XANN do not overload user’s cognitive capacity and most participants were able to use these explanations and apply the acquired information for a new task.
Vermeire et al.~\cite{https://doi.org/10.48550/arxiv.2107.04427}  proposed a methodology that receives information such as user requirements and needs, explanation purpose, stakeholder needs and constraints and based on the information suggest a ML model and explainability algorithm.
Another study focused on evaluating customer preferences for XAI in the Swedish Credit Scoring Industry dataset~\cite{matz2021explaining}. The findings revealed that customers prefer rule-based explanations. Three prototypes based on SHAP, LIME, and the counterfactual method were tested. Users were found to trust SHAP more than LIME and counterfactuals.

\begin{sidewaystable}[]
\centering
\caption{Summary of recent related works in explainble XAI.}
\begin{tabular}{|p{0.2\textwidth}|p{0.2\textwidth}|p{0.2\textwidth}|p{0.2\textwidth}|p{0.1\textwidth}|}
\hline
\textbf{General Category of Explanation} & \textbf{Paper Name} & \textbf{Main Contribution} & \textbf{Dataset} & \textbf{Is The Code Public?} \\
\hline
Feature Importance-based & Ribeiro et al.~\cite{ribeiro2016should} & Proposing Local Interpretable Model-agnostic Explanations (LIME) & 1- Books and DVDs dataset~\cite{blitzer2007biographies}, 2- Arbitarary images & \href{https://github.com/marcotcr/lime}{Yes} \\

\hline

Feature Importance-based & Lundberg et al.~\cite{lundberg2017unified} & Proposing SHAP (SHapley Additive exPlanations)& 1- MNIST digit dataset~\cite{deng2012mnist} & \href{https://shap.readthedocs.io/en/latest/index.html}{Yes} \\
		 \hline
  Feature Importance-based & Apley et al.~\cite{apley2020visualizing} & Proposing ALE (Accumulated Local Effect) & 1- Bike sharing rental counts~\cite{bike} & \href{https://pypi.org/project/PyALE/}{Yes} \\
		
		\hline
Counterfactual & Watcher et al.~\cite{wachter2017counterfactual} & Formulationg CFEs for the first time & No dataset & No  \\ 
		\hline

		Counterfactual & Ustun et al.~\cite{ustun2019actionable} & Using percentile as cost function + added feasibility to CFE & 1- Credit dataset~\cite{YEH20092473} & \href{https://github.com/ustunb/actionable-recourse}{Yes} \\ 
		\hline

		Counterfactual & Lucic et al.~\cite{lucic2019focus} &  Proposing an approach for non-differentiable tree ensembles & 1- Wine quality~\cite{dua2017uci}, 2- HELOC~\cite{FICO1}, 3- COMPAS~\cite{Ofer2017}, and 4- Shopping~\cite{dua2017uci}   & \href{https://github.com/a-lucic/focus}{Yes}   \\
		 
		\hline
		
		Counterfactual & Karimi et al.~\cite{karimi2020model} & Proposing an approach that is model-agnostic + actionable & 1- Loan approval (adult)~\cite{dua2017uci}, 2- Credit dataset~\cite{YEH20092473}, 3- Pretrial bail (COMPAS) dataset~\cite{Larson2016}& \href{https://github.com/amirhk/mace}{Yes}  \\		
  
		\hline
		
		Counterfactual & Karimi et al.~\cite{karimi2020algorithmic}  &  Proposing an approach based on Limited Knowledge of causality and gradient based optimization & No dataset& No \\
		\hline
		
		Counterfactual & Karimi et al.~\cite{karimi2021algorithmic} & Proposing an approach that shifted the focus from explanations to interventions & No dataset & No \\
		\hline
  
		Counterfactual & Mohammadi et al.~\cite{mohammadi2021scaling}& Proposing a framework based on Mixed-Integer Programming (MIP)  & 1- Loan approval (adult)~\cite{dua2017uci}, 2- COMPAS~\cite{Ofer2017}, 3- Credit dataset~\cite{YEH20092473} &\href{https://github.com/amirhk/mace}{Yes} \\		
		\hline

		Counterfactual & Mothilal et al.~\cite{mothilal2020explaining} & Proposing DICE algorithm that provides feasible + divers CFEs & 1- Adult income dataset~\cite{dua2017uci}, 2- Lending club dataset~\cite{Davenport2015}, 3- German credit dataset~\cite{dua2017uci}, 4- COMPAS dataset~\cite{Larson2016}	& \href{https://github.com/interpretml/DiCE}{Yes} \\
		\hline

		Counterfactual & Wang et al.~\cite{wang2021skyline} & Introducing the skyline (n space) for CFEs & 1- UCI adult dataset~\cite{dua2017uci},  2- Give Me Some Credit (GMSC)~\cite{creditfusion2011}, 3- HELOC dataset~\cite{explainablemlchallenge} & No \\
		\hline
	
		Counterfactual & Kanamori et al.~\cite{kanamori2021ordered} & Introducing a new objective function that evaluates a pair of actions and orders them based on feature interaction for CFEs & 1- FICO~\cite{FICO1}, 2- German credit dataset~\cite{dua2017uci}, 3- Wine quality~\cite{dua2017uci}, 4- Diabetes~\cite{dua2017uci} & \href{https://github.com/kelicht/ordce}{Yes} \\		
		\hline

		Counterfactual & Hada et al.~\cite{hada2021exploring} & Providing an algorithm to generate CFEs for regression trees & 1- Student Portuguese grades~\cite{dua2017uci}, 2- German credit~\cite{dua2017uci}, 3- Boston housing dataset~\cite{dua2017uci} & No \\		

		\hline
		
		Counterfactual & Verma et al.~\cite{verma2021amortized} & Proposing a stochastic control-based approach to generate sequential Algorithmic Recourses (AR)  &   1- German credit~\cite{dua2017uci}, 2- Adult income~\cite{dua2017uci}, 3- Credit default~\cite{dua2017uci} & No \\		
		\hline
\end{tabular}
\label{tab1}
\end{sidewaystable}
\clearpage

Counterfactuals showed the least system understanding, while SHAP was deemed the most useful. Cruz et al.~\cite{cruz2022evaluating} conducted an evaluation to determine user preferences between ``human-like" explanations and ``technical explanations". Their human-like explanations were derived from the probability of success in task completion by a robot, while technical explanations were generated based on Q-values in reinforcement learning. The authors found that non-expert participants consistently rated robot explanations focusing on the probability of success more favorably, with less variability, compared to technical explanations.

In 2020, Verma et al.~\cite{verma2020counterfactual} published a paper enumerating desired properties for counterfactual explanations, which are applicable to other XAI approaches as well.

\begin{enumerate}
\item 
    Actionability (plausibility): This property ensures that the CF explanation will refrain from advising changes to immutable features (e.g., race), focusing solely on mutable features (e.g., education, income).

\item 
    Sparsity: This property concerns the trade-off between the number of features changed and the total amount of change required to obtain the CFE. A CFE should alter a smaller number of features, aligning with the understanding that people find shorter explanations easier to comprehend~\cite{miller2019explanation}.
\item 
    Data manifold closeness: This property guarantees that CF explanations adhere to the data distribution, avoiding recommendations for changes that are unlikely to occur. For instance, if a patient is at high risk for heart disease, the system should not suggest losing 60 kg of weight in one year, as such a change is improbable.
\item 
    Causality: Features in a dataset are seldom independent; therefore, a change in one feature can impact others. For example, increasing job experience often correlates with increasing age. Thus, a counterfactual should preserve any known causal relations between features to be realistic and actionable.
\item 
    Amortized inference: Generating a counterfactual can be computationally expensive, requiring an optimization function for each data point. Amortized inference uses generative techniques, allowing the algorithm to compute a counterfactual (or several) for any new input without solving an optimization problem.
\item 
    Alternative methods: This property involves using guaranteed and optimal answers through mixed-integer programming or Satisfiability Modulo Theories (SMT) solvers. However, it imposes a limitation on classifiers with linear (or piece-wise linear) structures.
	
\end{enumerate}

In 2021, Verma et al.~\cite{verma2021counterfactual} identified obstacles impeding CFE deployment in the industry. In this work, they added two other desired CF properties:

\begin{enumerate}
	\item 
	Model-agnostic CFEs: An algorithm that can generate CFE for all black-box models, or in other words, a CF algorithm that is not restricted to a specific AI algorithm.
	
	\item
	Fast CFEs: This feature is about speed and is easy to use in practice. Fast CFEs refer to algorithms that can generate CFEs for multiple data points after a single optimization.
	
\end{enumerate}

The authors raised several concerns in the industry regarding CFEs. They emphasized that CFEs should not only be an interactive service but also explicitly indicate what should change and what should remain unchanged. Additionally, the authors argued that CFEs should be capable of capturing personal preferences.
While these properties are central to the CFE setting, it is crucial to note that they are subjects of ongoing debate. Furthermore, these properties can be evaluated and discussed in the context of other explanation methods. For instance, the model-agnostic property is not only preferred for CFEs but is also a desired quality for various other explanation approaches such as SHAP and LIME. The subsequent sections will delve into this concern in detail.

Brennen~\cite{brennen2020people} explored the expectations of various stakeholders regarding XAI and identified two primary concerns. Firstly, there is a need for consistent terminology in the field of XAI as different
disciplinary backgrounds developed different terminology. Secondly, there are multiple diverse use cases for XAI, such as debugging models, understanding bias, and building trust. Each use case involves different user categories, demanding distinct explanation strategies that are currently underdeveloped. They highlighted current XAI methods only explain to people with substantial expertise
in ML.
Alufaisan et al.~\cite{alufaisan2020does} compared the objective human decision accuracy without AI, with an AI prediction (without explanation), and with an AI prediction accompanied by an explanation. Their findings suggested that providing AI predictions enhances user decision accuracy. However, the evidence supporting the meaningful impact of explainable AI remains inconclusive. To draw this conclusion, the authors considered factors such as the accuracy of participants' decisions, confidence ratings, and reaction times.

The authors in~\cite{gerlings2020reviewing} mentioned the fundamental reasons we need XAI and in the following we introduce and discuss each one. 
\begin{enumerate}
    \item Generate trust, transparency and understanding: DARPA's XAI program~\cite{gunning2019darpa} underscores the importance of trust in XAI, emphasizing the need to comprehend and manage evolving AI systems. Gilpin et al.~\cite{gilpin2018explaining} stress the significance of models summarizing behavior to gain user trust and insight into decision-making. Miller~\cite{miller2019explanation} suggests two complementary approaches—interpretability and explanation—to create transparent systems. 

    \item Law requirement such as GDPR (General Data Protection Directive): in response to new regulations such as GDPR, there is a growing demand for XAI to not only provide explanations to users but also to society at large~\cite{arrieta2020explainable}. The introduction of GDPR, particularly the ``right to an explanation," has prompted considerable interest in XAI as a potential solution for regulatory compliance~\cite{goodman2017european},~\cite{kaminski2021}. Furthermore, some researchers advocate for the regulation of XAI itself or the establishment of standards to ensure responsible usage and prevent the development of persuasive yet opaque models~\cite{zerilli2019transparency}. 
   
    \item Social responsibility, fairness, and risk avoidance: particularly in healthcare, clinical practice, and justice systems, concerns over risks and accountability loom large, given the potential impact on human lives rather than just financial considerations~\cite{lipton2018mythos},~\cite{paez2019pragmatic}. The allocation of responsibility to individual professionals often leads to risk aversion. Therefore, there's a growing need to develop mental models for expert reasoning, such as clinical decision-making, to better understand the logic behind deep neural networks and other not transparent ML models~\cite{katuwal2016machine},~\cite{holzinger2017we}. Recent instances of discrimination and biased outcomes from opaque models have intensified discussions on ensuring fairness in model performance and gaining deeper insights into their construction. We will talk about these instances of discrimination in the following. 

    \item Generate accountable and reliable models: a significant draw towards XAI lies in its potential to ensure fairness and mitigate bias by allowing algorithm auditing and the creation of verifiable proof of their ethical soundness. Adadi and Berrada~\cite{adadi2018peeking} advocate for XAI as a means to audit algorithms, providing tangible evidence to defend algorithmic decisions as fair and ethical. Thus, the pursuit of algorithms that are not only fair and socially responsible but also accountable and capable of justifying their outputs further motivates the adoption of XAI. Regarding specific XAI methods, Doshi-Velez and Kim~\cite{doshi2017towards} argue for global explanations to understand entire models or groups scientifically, while local explanations are better suited for justifying specific decisions. 

    \item Minimize biases: biases in model performance serve as a significant catalyst for the adoption of XAI, spurred by media reports highlighting instances where models fall short compared to humans, such as in hiring processes or recognizing people of color~\cite{dastin2022amazon},~\cite{vincent2018google}.  Additionally, our cognitive biases can hinder the interpretation of model output, as we tend to oversimplify information~\cite{paez2019pragmatic}. 

    \item Being able to validate models: researchers have identified four evaluation methods for deep neural networks regarding biases and performance compared to humans: 1) completeness compared to the original model, 2) completeness on substitute tasks, 3) bias detection within models, and 4) human evaluations~\cite{gilpin2018explaining}. Evaluating XAI involves more than just precision and feature extraction, considering users' ability to understand model outputs. Some researchers proposed a comprehensive taxonomy for interpretability evaluation, including application-grounded, human-grounded, and functionally-grounded approaches. These methods range from testing implemented explanations with end-users to assessing broader quality notions without human interaction~\cite{doshi2017towards}.

\end{enumerate}

Chromik et al.~\cite{chromik2021think} conducted a study to understand how non-technical users form a global understanding of a model when provided with local explanations. Through empirical research, the authors sought answers to the following questions:
1- How robust is a self-reported global understanding gained from local explanations when examined? 2- How do non-technical XAI users construct a global understanding from local explanations?
The study utilized metrics such as accuracy of decisions, confidence, and perceived understanding to evaluate their findings. The authors reported a decrease in the perceived understanding of non-technical users when their perception was examined by researchers.
In 2020, Amann et al.~\cite{Amann2020} published a paper on the need for XAI in medicine, exploring perspectives from the legal, medical, patient, and ethical standpoints. Later, they investigated opinions on the explainability of AI-powered Clinical Decision Support Systems (CDSS)~\cite{Amann2022}. The review encompassed technical considerations, human factors, and the designated system's role in decision-making.

To sum up, there are three main key points that we can learn from reviewing works in the practice category:

\begin{enumerate}
	\item 
	
	 We can see that the most important properties for XAI that have been mentioned in the past works are as follow: actionability, sparsity, causality, data manifold closeness, and model-agnostic algorithms.

	\item 
	
	 There are specific needs and reasons to have XAI. Moral values (transparency and trust), debugging, eliminating bias, model validation, intervention and changing the output, and law requirement (GDPR).

	\item 
	
	 Users of XAI can be classified into three groups: stakeholders, expert users, and non-expert users.
\end{enumerate}

\section{Who are users of XAI, what they want from XAI, and how deliver it: a Mapping Function}
\label{map}

After reviewing proposed methods in XAI and their potential applications in Section~\ref{pre}, this section aims at addressing key questions regarding XAI: Who are the users of XAI? What are their purposes for utilizing XAI? What properties are beneficial for each user and goal? Lastly, which methods are suitable for different goals? We answer the mentioned questions by proposing a mapping function that consider user's needs, their goals, and input data, and based on that information can suggest the potential XAI methods and properties. The proposed mapping function in this section assesst all users of XAI in selecting the most suitable methods and properties for their specific purposes. The ultimate goal is to refine XAI methods based on the tasks they are supposed to perform, ensuring that they are chosen carefully to enhance their practical effectiveness. The outcome of this section is expected to contribute to the development of XAI methods that are more aligned with users' objectives and exhibit improved performance in real-world applications.

The significance of this lies in the recognition that researchers often overlook the ultimate goal of each explanation and the specific properties that are useful for achieving those goals. Some XAI properties have been proposed without adequate consideration of whether each characteristic is beneficial for a particular purpose, such as sparsity~\cite{van2022explainable}. The objective here is to delve into the users of XAI, understand their motivations for applying XAI, and identify potential properties of XAI that align with each user's goals. In the following we gathered all the necessary elements to build the proposed mapping function.

In Section~\ref{practice}, it has been comprehensively discussed that XAI has been used for the following purposes~\cite{gerlings2020reviewing}:
1- Generating trust and transparency, 2- XAI is turning to customers' rights (e.g., general data protection regulation law), 3- Fairness and risk avoidance, 4- Generating accountable models, 5- Minimizing biases, 6- System Validation.

In the case of users, we have three categories of users~\cite{brennen2020people}:

\begin{enumerate}
    \item Stakeholders such as banks, hospitals, and institutions.
    \item Experts that can be AI experts such as developers and non- AI domain experts such as doctors.
    \item Not expert users, such as an applicant for loan applications or patients.
\end{enumerate}  
In the next section, we will discuss each category of AI users in details.

The other element of this mapping function that we discussed in Section~\ref{practice} is the desired properties for XAI which were as follow~\cite{verma2020counterfactual},~\cite{verma2021counterfactual}:
1- Plausibility (actionability), 2- Sparsity, 3- Run time, 4- Being model agnostic, 5- Data manifold closeness, 6- Causality, 7-Fairness.

In the case of approaches to provide explanations, we considered the following developed approaches: counterfactual explanations~\cite{wachter2017counterfactual},~\cite{karimi2020algorithmic}, Local Interpretable Model-agnostic Explanations (LIME)~\cite{ribeiro2016should}, Shapley Additive exPlanations (SHAP)~\cite{lundberg2017unified}, saliency maps~\cite{simonyan2013deep}, Partial Dependence Plot (PDP)~\cite{friedman2001greedy}, Individual Conditional Expectation (ICE)~\cite{goldstein2015peeking}, Accumulated Local Effects(ALE)~\cite{apley2020visualizing}, Layer-wise relevance propagation (LRP)~\cite{bach2015pixel}, Class Activation Mapping (CAM)~\cite{zhou2016learning}, and Gradient-weighted Class Activation Mapping (Grad-CAM)~\cite{selvaraju2017grad}.

In general, we have three ways to present explanations: 
\begin{enumerate}
    \item Visual explanations such as plots.
    \item Textual explanations such as text.
    \item Mathematical or numerical explanations.
\end{enumerate}

Explanations can be provided for various purposes such as classification, regression, and segmentation tasks.

Finally, the last element of the mapping function is the format of data. Data that has been used in AI algorithms can be tabular or numerical data, image data, and text data. For example, some explainability approaches are proposed for Artificial Neural Networks (ANNs) that are mainly used for image data.

\begin{figure}
    \centering
    \includegraphics[scale=0.5]{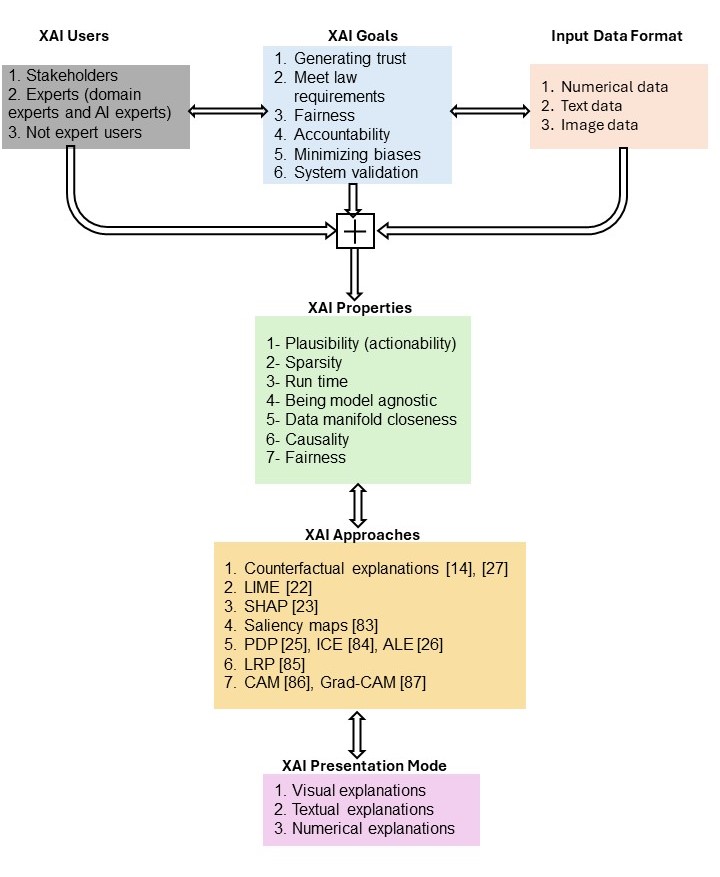}
    \caption{Taxonomy of XAI users, their motivations, types of data, desired properties, XAI approaches, and methods of XAI presentation. By knowing who the users of XAI are plus their goals for using XAI plus data format as inputs, we can map them to an appropriate set of properties, XAI methods, and presentation modes.}
    \label{highlevel}
\end{figure}

Fig.~\ref{highlevel} demonstrates what we discussed earlier and depicts the interaction of different parts. As the figure shows, XAI users, their motivation to implement XAI, and the type of data that they are dealing with are connected to each other and can affect the choice of related XAI desired properties. All these four parameters—users, their motivation, the type of data, and desired properties—will help us choose the proper XAI algorithm and method of presentation of explanations.

\subsection{Who Wants XAI?}

In this section, we delve into categorizing XAI users and provide some real-world examples. We categorized users to three categories of stakeholders, domain expert users, and non-expert users.

\subsubsection{Stakeholders}
The first category of users comprises stakeholders, referring to individuals or organizations that are not experts in AI but are implementing AI to provide a service. They can include policymakers, institutions, banks, employees, investors, etc~\cite{brennen2020people}. These stakeholders benefit from XAI by using it for moral values, validating the systems provided by engineers, meeting legal requirements, and gaining the trust of their users.

However, stakeholders are not directly involved in fixing the system, eliminating bias, or changing its output, as they do not possess sufficient knowledge to do so. An example of a stakeholder is a company hiring people. By providing explanations to applicants who were not hired, the company demonstrates its commitment to transparency and fairness while also gaining the trust of its applicants. One critical consideration in this case is tailoring the explanations in order to meet stakeholders needs~\cite{kim2024stakeholder},~\cite{hoffman2023explainable}.

\subsubsection{Domain Expert Users}

Domain expert users, also known simply as expert users, constitute another category of AI users who leverage XAI to achieve specific goals. These experts fall into two primary categories: AI experts and non-AI domain experts. AI experts encompass individuals such as developers and researchers. Although AI experts may not be directly influenced by AI, they bear the responsibility of its development, debugging, and validation. On the other hand, non-AI domain experts include professionals like doctors, pathologists, and police officers. While these experts may not directly engage with AI technologies, they rely on AI to inform decisions or execute tasks.
Both AI experts and non-AI domain experts can employ XAI to verify or validate systems, ensuring alignment with ethical and legal standards. For instance, imagine an XAI algorithm providing insights into the output of a medical diagnosis AI system. Here, AI experts would utilize XAI to examine the algorithmic aspects, while doctors would leverage it to evaluate the medical components. This example underscores the necessity for collaboration between these two groups to achieve specific objectives, such as debugging. Furthermore, both groups can utilize XAI to foster trust in the AI systems they develop (AI experts) or utilize (non-AI domain experts)~\cite{bayer2022role}.
Neither AI experts nor non-AI domain experts typically aim to alter AI-based decisions, as these decisions usually lie outside their direct purview. For example, a doctor might employ counterfactual explanations to understand why a patient is classified as low-risk rather than high-risk for heart disease. In this scenario, the doctor utilizes the explanation as a means to comprehend the patient's situation, rather than for personal gain.

\subsubsection{Non-expert Users}

The last category of XAI customers consists of non-expert users. These individuals are not experts in AI or any other field combined with AI, but they directly experience the impact of AI decisions. Examples of non-expert users include patients in hospitals, loan applicants, and job seekers. Non-expert XAI users aim to intervene and modify the outputs generated by AI algorithms, exercise the rights granted by the law, and place trust in the systems' trustworthiness. These users lack the position to debug the system or eliminate bias and do not possess sufficient knowledge to validate the system~\cite{gilpin2022explanation}. This category of users is considered more vulnerable compared to others, as they can easily be misled or confused by XAI and have limited information about AI.

To illustrate this vulnerability, consider an AI system that classifies patients into two groups, those in immediate need of cardiac surgery and those who do not require it. If the algorithm erroneously classifies a patient in an emergency situation as someone who does not need surgery, both the classification and its corresponding explanation will be incorrect. Now, envision two groups of people exposed to this explanation: doctors and patients. Doctors, equipped with medical knowledge, are more immune to false information provided by explanations. However, this misleading XAI information can lead to irreversible harm to patients. In conclusion, the level of sophistication in the explanation should be modified based on the user's expertise level~\cite{doshi2017towards}.

\subsection{What Properties Should XAI Have?}

So far, we introduced XAI users and established a clear understanding of why each user of XAI utilizes it. In this section we discuss the desired properties for different goals such as system validation, bias elimination, law requirement, moral values, trustworthiness, system debugging, and output alternation. 

\begin{itemize}
    \item Plausibility: plausibility~\cite{verma2020counterfactual} is the most fundamental property that renders an explanation valid. It is a necessary condition for all purposes of XAI. It is challenging to envision people placing trust in AI systems that provide recommendations that are not plausible or actionable.

    \item  Model-agnostic: this property is applicable to all XAI purposes. Validating systems using XAI methods designed for various ML algorithms proves to be time-saving and more trustworthy. This property ensures that XAI explanations are not confined to specific models, enhancing their applicability and reliability across a range of machine learning approaches~\cite{verma2021counterfactual}.

    \item Low run-time: If XAI is utilized for purposes such as system validation, debugging, and bias elimination, run-time becomes less critical. In these cases, XAI is not employed as a real-time task for the mentioned purposes. Therefore, time should not be a significant factor in this trade off. It is essential to clarify that when stating run time is not an issue, it does not imply an infinite run-time, as in the case of some brute force algorithms. In these scenarios, superior performance is prioritized over low run-time~\cite{verma2021counterfactual}. However, when non-expert users are using XAI for intervention purposes, run-time becomes a crucial factor as faster systems are more desirable. Moral values favor explanations that take less time. Although there is no clear study establishing a direct connection between trustworthiness and run-time in XAI systems, in general, providing explanations more quickly tends to be more successful in earning trust. It is noteworthy that current legal requirements, such as the GDPR law, focus on the clarity and accessibility of information without explicit reference to the run-time of explanations. For instance, the GDPR law emphasizes that ``information must be concise, transparent, intelligible and easily accessible, and use clear and plain language"~\cite{eu-269-2014}. As seen, there is no mention of the run-time of explanations. However, establishing a threshold for the required time for generating explanations could be a consideration.

    \item Sparsity: with the same logic mentioned for the low run-time property, we can assert that sparsity is not an on-demand property for debugging, validating, and eliminating bias in XAI systems. We can envision scenarios where longer sentences can provide more information about algorithm performance. For instance, in a counterfactual explanation, a recommendation suggests changing ten features in a patient to classify them as a low-risk heart disease patient rather than a high-risk patient. Eight out of these ten features are related to the patient's job life, such as reducing working hours and stress levels. With this lengthy explanation that does not adhere to the sparsity property, we can argue that the system has over-weighted the importance of patients' job features instead of suggesting changes to variables like weight, eating habits, etc. For intervention purposes, sparsity is a desirable property. In general, sparsity is a preferred characteristic when a non-expert user is involved; otherwise, it is not a critical feature. Since shorter explanations are more understandable, they garner more trust. Similar to low run time, there is no legal requirement regarding sparsity, but it can be implemented as a requirement.

    \item Causality: causality and correlation between features are important properties that many XAI methods currently lack~\cite{verma2020counterfactual}. This attribute is crucial for altering the output of AI systems and for the comprehensive validation of the system. XAI should be able to explain the causality between features to ensure that the AI algorithm interprets it correctly. While other validation methods, such as checking for bias, can be useful, examining causality is essential for complete validation. Therefore, the ability to explain causality is a desirable property for all AI applications.

    \item Data manifold closeness: in the context of detecting and eliminating bias in AI systems, being loyal to the data distribution (data manifold closeness) is desirable, especially for objectives such as intervention or moral values. However, it might not be the case in every scenarios. For example, in a dataset where the majority of people hold a B.Sc. degree, if the goal is to detect bias for M.Sc. degrees, the desired explanations should focus on the M.Sc. degree and reveal the AI algorithm's strategy for this specific degree. In this case, being loyal to the data distribution is not the appropriate property.
\end{itemize} 

In Table~\ref{property_mix}, we have summarized all XAI users and the potential purposes for using XAI and assigned desired properties for users and their goals.

\subsection{Which XAI Method Should We Use to Meet Our Goals?}

So far, we have associated different users with distinct goals and linked various purposes to specific properties. In this section, we map different XAI generation methods to these goals and desired properties.

\subsubsection{When We Use Counterfactual Explanations?}

As previously discussed, CFE offer a framework for suggesting changes to achieve a desired outcome. Within this paradigm, such methods can be effectively utilized by AI experts, non-AI domain experts, and non-expert users seeking to intervene and modify outputs. Notably, while non-expert users may employ CFEs to directly influence outcomes pertinent to them, expert users would utilize CFEs to impact outcomes relevant to non-expert users. Although CFEs initially proposed as a means of explainability for non-expert users, with less emphasis on goals such as system validation, debugging, and bias elimination, recent years have seen researchers leverage the concept of ``counterfactuals" to develop algorithms catering to these needs.

Abid et al.~\cite{abid2022meaningfully} proposed Conceptual Counterfactual Explanations (CCE), which combines counterfactual explanations with concept activation vectors, elucidating why a sample might be misclassified and what changes are required for correct classification. Their algorithm, tested on both natural and medical images, demonstrated efficacy. However, not testing CCE's applicability beyond image data and the lack of user studies supporting their results remain notable gaps.
In another endeavor, Iqbal et al.~\cite{iqbal2021cadet} introduced CADET (Causal Debugging Toolkit), employing causal path analysis and counterfactual reasoning to identify fault origins, assess configuration impacts, and propose fixes for computing platforms. CADET exhibited superior accuracy compared to prior works and proved effective in real-world fault examinations, outperforming expert advice within a notably shorter timeframe.
Similarly, Goyal et al.~\cite{goyal2019counterfactual} utilized counterfactual reasoning to generate counterfactual visual explanations in order to debug the system and find the necessary changes that need to be applied on image $I$ so the model predicts it as class $c^{\prime}$ instead of $c$~\cite{goyal2019counterfactual}. The authors tested their method on Caltech-UCSD Birds (CUB) 2011 dataset~\cite{wah2011caltech} and they reported on average it takes 7.4 edits and  1.85 sec/image to change the model’s prediction from $c$ to $c^{\prime}$.

\begin{table}[ht]
	\centering
	\caption{XAI Users, their purposes for using XAI, and desired properties for them.}
	\begin{tabular}{ | l | l| l|l|} 
	\hline
        \diagbox[width=10em]{Goals}{Users} & Stakeholders& Expert Users & Non-expert Users  \\
		\hline
		
		& Plausibility & Plausibility & \\
        System Validation & Model- agnostic & Model- agnostic & \xmark \\
         & Causality & Causality & \\
         & Data Manifold Closeness & Data Manifold Closeness & \\
  \hline
  &  & Plausibility &\\
  &  & Model- agnostic &\\
  System Debugging & \xmark & Causality & \xmark\\
   &  & Data Manifold Closeness&\\

  \hline
		  &  & Plausibility &\\
   Eliminate Bias & \xmark & Model- agnostic & \xmark\\
   &  & Causality &\\

   \hline

       		   &  &  & Plausibility \\
         &  &  & Model- agnostic   \\
         &  &  & Sparsity  \\
         System Intervention & \xmark & \xmark & Causality  \\
         &  & & Run Time \\
         & &  & Data Manifold Closeness \\

\hline

		   & Plausibility & Plausibility & Plausibility \\
         & Model- agnostic & Model- agnostic & Model- agnostic  \\
         & Sparsity & Sparsity & Sparsity  \\
        Moral Values & Causality & Causality &  Causality \\
         & Run Time & Run Time & Run Time \\
         & Data Manifold Closeness &  Data Manifold Closeness &  Data Manifold Closeness \\

  \hline

  		   & Plausibility & Plausibility &  Plausibility\\
         & Model- agnostic & Model- agnostic & Model- agnostic \\
         & Sparsity & Sparsity & Sparsity  \\
        Trustworthiness & Causality & Causality & Causality  \\
         & Run Time & Run Time & Run Time \\
         & Data Manifold Closeness &  Data Manifold Closeness &   Data Manifold Closeness\\

  \hline

    		   & Plausibility & Plausibility & Plausibility \\
         & Model- agnostic & Model- agnostic & Model- agnostic   \\
         & Sparsity & Sparsity & Sparsity  \\
        Law Requirement & Causality & Causality & Causality  \\
         & Run Time & Run Time & Run Time \\
         & Data Manifold Closeness & Data Manifold Closeness & Data Manifold Closeness \\

	\hline
	
	\end{tabular}
	\label{property_mix}
\end{table}

\textbf{Output summary:} while CFEs primarily role is to serve to non-expert users by providing actionable insights for outcome alteration, experts—both AI and non-AI domain experts—can leverage counterfactual reasoning for system debugging and validation. Essential properties such as plausibility, model-agnosticism, and low runtime, crucial for non-expert users, retain significance from a desired properties perspective.

\subsubsection{When We Use Local Interpretable Model-agnostic Explanations (LIME)?}

Many XAI approaches such as saliency map~\cite{simonyan2013deep}, LIME~\cite{ribeiro2016should}, and SHAP~\cite{lundberg2017unified} rely on the concept of feature importance. The fundamental idea behind feature importance in XAI is to assess the contribution of each feature to a prediction and justify the output based on these contributions. Features in these approaches can take various forms, including numerical values in tabular data, pixels in images, and words in text data.

Local Interpretable Model-agnostic Explanations (LIME)~\cite{ribeiro2016should} is a method specifically designed to provide local explanations for nonlinear AI algorithms. Local explanations aim to explain the predictions of a model for individual instances or data points. LIME achieves this by constructing a linear model that approximates a selected instance within the nonlinear algorithm, making the linear model more interpretable. This approach proves valuable for validating and debugging AI systems, mitigating bias, and enhancing transparency in decision-making processes. LIME effectively communicates the rationale behind an algorithm's decision to stakeholders, experts and non-experts, thereby fostering trust in the system.
LIME can be considered a user-friendly XAI method, especially for non-AI domain experts and non-expert users, as both groups may lack AI backgrounds, making user-friendliness crucial. For instance, one study evaluated how explanations provided by LIME compared with those independently provided by physicians~\cite{kumarakulasinghe2020evaluating}. The findings indicated that LIME explanations were clinically relevant and highly concordant with explanations provided by physicians, with clinicians expressing high levels of trust and reliance on LIME. Another study conducted a user study using the German Credit dataset~\cite{dua2017uci} to assess how non-expert end-users perceive and evaluate XAI methods like LIME~\cite{meza2023does}. Participants described LIME as easy to understand and visualize, more intuitive and simple than other approaches such as SHAP. However, concerns have been raised about LIME's substantial amount of randomness, which results in a lack of robustness~\cite{zhang2019should}. Additionally, it has been argued that the perturbed samples can contain many incorrectly classified instances, as the sampling process does not consider the density of the data~\cite{guidotti2019factual}, which can produce data points with high prediction uncertainty.
Alvarez-Melis and Jaakkola~\cite{alvarez2018robustness} argue that similar inputs should yield similar explanations and introduce a metric, local Lipschitz estimate, to measure the stability of XAI methods like LIME and as the metric is lower, the explanation is more robust. They found that, in some cases, despite minimal changes in the classifier's predicted class probability due to perturbations, LIME explanations varied considerably.

\textbf{Output summary:} although LIME can be used by all categories of XAI; it needs to be under supervision of AI-experts to control and validate provided explanations.
Regarding intervention, LIME provides a simplified decision boundary, indicating that modifying the input to fall on the opposite side of the estimated boundary may alter the output. However, it does not specify which features to change or by how much, making it less suitable for system intervention. Concepts such as plausibility, sparsity, and data manifold closeness are not applicable in this context. Desirably, LIME should have a low runtime and be model-agnostic. The next desired property is causality that LIME does not consider it. To address this issue one work proposed a new method named BayLIME based on LIME that use prior knowledge and Bayesian reasoning to improve causality (also named fidelity)~\cite{zhao2021baylime}.

\subsubsection{When We Use SHapley Additive exPlanations (SHAP)?}

SHAP (SHapley Additive exPlanations) is a visualization method~\cite{lundberg2017unified} that illuminates the contribution of each feature to the output for a specific instance. Meza Martínez et al.~\cite{meza2023does} examined how non-expert users interact with SHAP, utilizing the German Credit dataset~\cite{dua2017uci} for their evaluation. Their findings revealed that non-expert users perceived SHAP as complex and hard to understand, primarily due to its handling of probabilities. However, upon comprehension, participants found SHAP to be informative. Adding this results to Cruz et al.'s~\cite{cruz2022evaluating} study, which favored human-like explanations over probability-based ones, both studies suggest that non-expert users struggle to connect with probability-based explanations, regardless of format—whether image or text.
SHAP offers the capability to calculate both local and global explanations. Global explanations delineate the overarching behavior of a machine learning model across the entire dataset, as opposed to individual data points. While Shapley values theoretically ensure locally accurate and replicable explanations, their exact computation proves computationally intensive~\cite{meza2023does}. Alvarez-Melis and Jaakkola~\cite{alvarez2018robustness} observed that while both LIME and SHAP yield stable explanations for linear models, their consistency diminishes significantly within small neighborhoods for nonlinear models. In general, SHAP tends to exhibit more stability compared to LIME, albeit subject to variability depending on the dataset.

The utility of SHAP extends to system validation, debugging, and bias mitigation, allowing experts to gauge whether specific features significantly influence model output. For instance, when assessing explanations for a rejected job applicant, SHAP facilitates examination of whether the applicant's gender or race substantially impacts the outcome. Gwak et al.~\cite{gwak2023debugging} utilized SHAP for debugging malware classification models in cybersecurity, while Hickey et al.~\cite{hickey2021fairness} sought to mitigate model bias using regularization derived from SHAP values of an adversarial surrogate model.
Though SHAP inherently does not depict feature correlation and causality, efforts have been made to extend its capabilities to address feature dependency and causal structures~\cite{carballo2022explainability},~\cite{heskes2020causal}. Overall, intervention remains an unviable option with this method, while low runtime and model-agnosticism emerge as expected properties.

\textbf{Output summary:} SHAP emerges as a computationally demanding yet insightful explainability method, better suited for AI experts or non-experts with a foundational understanding of AI, rather than novices in the field.

\subsubsection{When We Use Saliency Map?}

Saliency maps were initially introduced as a visualization tool for image classification models rather than an explicit method within XAI~\cite{simonyan2013deep}. These maps visualize pixels that are more important for the ML algorithm and are typically applied to image datasets. 
Researchers have found that saliency maps can aid users in understanding specific image features the system is sensitive to, thereby enhancing their ability to predict the outcome of the network for new images~\cite{alqaraawi2020evaluating}. However, even with saliency maps available, Convolutional Neural Networks (CNNs) often remained largely unpredictable for participants.
Studies by Kindermans et al.~\cite{kindermans2019reliability} have indicated that most saliency methods lack invariance under simple input transformations and are highly sensitive to the choice of reference point. Melis and Jaakkola~\cite{alvarez2018robustness} provided an example where a perturbed version of an image, virtually indistinguishable from the original to the human eye, yielded remarkably different explanations. Nevertheless, they also noted that saliency maps demonstrated better stability compared to LIME and SHAP methods.

Despite their utility, saliency maps have limitations. They do not elucidate how the network arrived at a specific decision, hence lacking sufficient information for intervention to alter the output. Nonetheless, saliency maps find applications in system validation, debugging, bias mitigation, establishing trust, and meeting legal requirements~\cite{alsallakh2021debugging},~\cite{bertrand2022saliency},~\cite{ayhan2022clinical}.
Saliency maps do not directly address plausibility, sparsity, or data manifold closeness, and model-agnosticism and low runtime are desired properties. In terms of causality, while image data inherently possess a connectivity matrix showing pixel correlations, saliency maps do not visualize this matrix. Consequently, computed saliency maps may not always reflect intended saliency and could be independent of both the model and the data generating process~\cite{adebayo2018sanity},~\cite{lapuschkin2019unmasking}. To mitigate this, the CHIME framework has been proposed~\cite{biswas2022chime}, which begins by identifying salient patches of the input that the model prioritizes and then employs causal discovery techniques to unveil causal relationships among various concepts.

\textbf{Output summary:} saliency maps offer a straightforward and intuitive visualization of influential parts of input data, primarily used for images, aiding in understanding the model's behavior. However, they can be sensitive to input perturbations or noise, potentially leading to unstable explanations. Thus, they can be used by non-expert users; however, require AI-expert oversee.

\subsubsection{When We Use Partial Dependence Plots (PDPs), Individual Conditional Expectation (ICE), and Accumulated Local Effect (ALE)?}

Partial Dependence Plots (PDPs) serve as a method for evaluating feature importance in AI algorithms and providing global explanations~\cite{friedman2001greedy}. PDPs illustrate the contribution of each feature in the decision-making process by assuming that features are independent. Krause et al.~\cite{krause2016interacting} developed an interactive partial dependence diagnostics to help AI experts understand how features affect predictions overall and allows them to understand how and why specific data points are predicted as they are. They reported that through extended use of the proposed method, they achieved better predictive models, as well as improved communication of models to related stakeholders.

While PDPs provide valuable insights into how individual features impact model predictions, they do not infer causality, making them unsuitable for effecting changes to the output and altering them. Some research argues that it is possible to extract causal information from PDP and ICE plots, but this does not come for free~\cite{zhao2021causal}. In summary, the following requirements have been mentioned to have a successful causal interpretation: 1) A good predictive model, so the estimated black-box function is close to the real one. 2) Some domain knowledge about the causal structure to assure the necessary condition is satisfied. 3) Visualization tools such as the PDP and its extension ICE. Loftus et al.~\cite{loftus2023causal} developed Causal Dependence Plots (CDPs) based on PDP that can visualize how one variable, an outcome, depends on changes in another variable, a predictor, along with any consequent causal changes in other predictor variables, addressing the issue of PDPs assuming features are independent. In terms of robustness, PDPs' robustness depends on factors such as ML model complexity, data quality, and ML model robustness. However, PDPs are not immune and can be manipulated by changing the data~\cite{baniecki2022fooling}.

PDPs enable us to observe how predictions change by modifying a single feature, aiding in the assessment of whether the system is using the correct features and controlling bias. Although PDPs can be helpful for bias elimination, they carry the risk of generating biased interpretations~\cite{moosbauer2021explaining}. Plausibility, sparsity, low runtime, and being model-agnostic are desired properties (as we mentioned before these are desired properties for most XAI methods). While this method does not explicitly consider causality, efforts have been made to enable it to consider causality~\cite{loftus2023causal},~\cite{moosbauer2021explaining} by showing how the model output depends on changes in a given predictor along with any consequent causal changes in other predictors.

Regarding closeness to manifold data, this method considers all features constant except one, varying that feature across a wide range. For some values within this range, the feature may be unlikely or not close to the dataset distribution center. For instance, considering the effect of education feature among job applicants, PDPs may include high school education, B.Sc., M.Sc., and Ph.D., even though Ph.D. and high school are not close to the center of the education feature distribution.

Individual Conditional Expectation (ICE) illustrates how the output changes when modifying a feature within an instance in the input set~\cite{goldstein2015peeking}. ICE plots refine the PDP by depicting the functional relationship between the predicted response and the feature for individual observations. Unlike partial dependence plots, which display one line overall, an ICE plot presents the dependency of the prediction on a feature for each instance, resulting in one line per instance~\cite{molnar2020interpretable},~\cite{huang2023beyond}. However, ICE has its limitations. Firstly, it can effectively plot only one feature per plot, as more features require more surfaces that would overly and complicate the visualization. Secondly, if the feature being studied is correlated with other features, certain points in the lines may become invalid data points according to the joint feature distribution, similar to the issue encountered with PDP. Lastly, as the number of features or instances increases, the plots may become overcrowded, or numerous plots would need to be generated, making interpretation challenging~\cite{molnar2020interpretable}. To address these challenges, ALE has been introduced.

Accumulated Local Effect (ALE) is another model-agnostic approach based on feature importance that provides global explanations and is closely related to PDP~\cite{apley2020visualizing}. Essentially, this method plots the influence of a feature on the algorithm's prediction, taking into account the feature's correlations, unlike PDPs, and is less computationally expensive compared to PDP. Gkolemis et al.~\cite{gkolemis2023dale} mentioned that ALE does not scale well in cases where the input has high dimensionality which makes it vulnerable to out-of-distribution (OOD) sampling when the training set is relatively small. Okoli~\cite{okoli2023statistical} mentioned three challenges with conducting statistical inference based on ALE: it is hard to verify reliability of ALE analyses, especially in the context of small datasets; it is challenging to intuitively characterize a variable's overall effect in ML; and making robust inferences from ML data analysis can be demanding.

Based on the discussed topics, we can conclude that stakeholders and experts can effectively use the ALE approach. Its visual nature makes it usable for non-experts as well. ALE plots are preferred over PDPs for several reasons: they eliminate unreliable bias cases in the PDP, produce optimal results despite feature correlations, and are less computationally expensive~\cite{dwivedi2023explainable}. However, Huang et al.~\cite{huang2022analysis} cautioned that ALE might not always be the default better choice. If AI-expert users cannot determine the proper order of the categorical feature of the target model, they cannot explain this feature with ALE since ALE can only be executed on the ordered feature space. Furthermore, as ALE does not consider causality, it is unsuitable for intervention and changing output purposes. Plausibility, sparsity, low runtime, and model-agnosticism are desirable properties for ALE. As previously mentioned, these attributes are also sought after in most XAI methods. Regarding closeness to manifold data, the same discussion as in PDP applies here. In ALE, the algorithm considers all values in the dataset for a feature, and some of them may be disloyal to the data distribution.

\textbf{Output summary:} the most significant drawbacks of PDPs are that they consider features to be independent (lack of causality and correlation), the explanation quality depends on the trained ML model as the explanation is using the ML model to generate plots, and PDPs can also be misinterpreted if users do not carefully consider the underlying assumptions and limitations, which becomes more dangerous for non-expert users. However, PDPs can present better intuitive information to both AI experts and non-experts compared to ALE and visualize how features are interacting with each other for system validation and bias elimination. On the other hand, ALE considers correlation between features and has compared less computational complexity but interpretation of plots might be more challenging for users without AI background.

\subsubsection{When We Use Layer-Wise Relevance Propagation (LRP)?}

Layer-Wise Relevance Propagation (LRP), introduced by Bach et al.~\cite{bach2015pixel}, is a model-agnostic explanation method designed to elucidate how neural networks operate. This technique is grounded in the backward propagation of a prediction to ascertain the relevance of each feature to the predicted output. The output of this method is a heatmap indicating the relevance and contribution of each pixel to the predicted output. From a stability perspective, LRP stands somewhere in between; it is less stable than saliency maps and more stable than LIME~\cite{alvarez2018robustness}. Although LRP can be used for images, videos, or text data, it has mainly been applied to image data.
Huber et al.~\cite{huber2019enhancing} argue that LRP highlights every possible cause for a specific prediction on its saliency map, leading to attention being dispersed across diverse parts of the input. This can make the results challenging to interpret for non-expert users. They proposed an adjustment to the LRP concept that utilizes only the most relevant neurons of each convolutional layer by employing an $argmax$ function to find the most contributing neurons. This adjustment generates more restricted saliency maps with fewer highlighted areas.
Researchers have contended that LRP only furnishes partial insights, placing the burden of interpreting the model’s reasoning to the user~\cite{achtibat2023attribution}. Consequently, they introduced Concept Relevance Propagation (CRP), which, while built upon LRP, provides explanations that are more readily understood by humans. CRP disentangles the relevance flows linked to the concepts learned by the model through conditional backpropagation. This enables the computation of concept-conditional relevance maps $R(x|\theta)$, where $x$ denotes the data point predicted by the model, $\theta$ describes a set of conditions, and $R$ signifies the relevance quantity.
Samek et al.~\cite{samek2016evaluating} discussed that while the usefulness of heatmaps provided by approaches such as LRP can be subjectively judged by a human, an objective quality measure is missing. They presented a methodology based on region perturbation for evaluating ordered collections of pixels such as heatmaps. They reported that the LRP algorithm qualitatively and quantitatively provides a better explanation of what led a Deep Neural Networks (DNNs) to a particular classification decision than other approaches such as the sensitivity-based approach (saliency map)~\cite{simonyan2013deep} or the deconvolution method~\cite{zeiler2014visualizing}. They attribute this performance to the fact that heatmaps computed with the deconvolution algorithm and sensitivity-based algorithms are much noisier than those computed with the LRP method, making them less suitable for identifying the most important regions with respect to the classification task and for debugging purposes.

Sundararajan et al.~\cite{sundararajan2017axiomatic} explain that LRP addresses the sensitivity problem by using a starting point or baseline and computes ``discrete gradients" instead of (instantaeneous) gradients at the input. Yet, LRP swaps out regular gradients with these discrete ones, but still relies on a changed form of backpropagation to combine them into explanations. Because discrete gradients do not follow the chain rule, LRP fails to satisfy implementation invariance.
Ullah et al.~\cite{ullah2021explaining} compared LRP's performance with other approaches such as SHAP and LIME for tabular data on Credit Card Fraud detection~\cite{ulb2021creditcard} and Telecom Customer Churn prediction~\cite{bandi2019telecom} datasets using a one-dimensional CNN. They reported that LRP achieved better computational performance (LRP: 1–2 seconds, LIME: 22 seconds, and SHAP: 108 seconds), making it unique for real-time applications. Additionally, LRP can highlight features for enhancing the model performance, thus aiding in feature subset selection.

Researchers have applied LRP in bias elimination scenarios. For instance, Seetharam~\cite{seetharam2023u} used LRP to show that although the U-Net convolutional neural network~\cite{ronneberger2015u} has high accuracy, it is hindered by a bias to prioritize color over shape in imaging-based models. The researcher suggests that LRP results should be used to ensure that the training set is not biased, and U-Net would need to be modified before being applied to colored images.
Although LRP demonstrates the most important regions for decision-making, it does not reveal how these regions affect the final outcome and may still leave some aspects of the model's behavior unexplained. 
As this approach does not clarify the correlation and causality between pixels, also known as connectivity matrix, it is not suitable for intervention and altering the output. 

\textbf{Output summary:} LRP can offer well-detailed and fairly stable and fast explanations regarding the performance of neural networks. However, interpretation of information provided by LRPs can be challenging for people with limited AI background and this approach does not clarify how the system made a decision. It can be used by AI users for a wide range of purposes such as bias elimination and system validation. However, it needs to be modified before being represented to non-AI expert users.

\subsubsection{When We Use Class Activation Map (CAM) and Gradient-weighted Class Activation Mapping (Grad-CAM)?}

Class Activation Map (CAM)~\cite{zhou2016learning} offers an additional avenue for explicating CNNs, particularly in image contexts, by highlighting pivotal pixels that have significant influence over the CNN's classification output. This method employs global average pooling functions coupled with class activation mapping to pinpoint the locations of class-specific image regions within an input image. CAM operates by modifying the CNN architecture, substituting the fully connected layer with a Global Average Pooling layer (GAP). Within this framework, the GAP layer computes the average contribution of each feature map in the final convolutional layer, followed by a weighted summation of vectorized averages to generate the activation map. Subsequently, the CAM model overlays this activation map onto the input image, thereby identifying the areas of interest exploited by the CNN in making its predictions. A notable drawback of CAM stems from its architectural modifications, which can impact prediction accuracy~\cite{ibrahim2023explainable}. Consequently, researchers proposed the Gradient-weighted Class Activation Mapping (Grad-CAM) model~\cite{selvaraju2017grad}. Unlike CAM, Grad-CAM avoid altering the architecture and instead computes gradients of a predicted class in the last convolutional layer. It selectively considers features that positively influence class prediction while consider negative features irrelevant to the classification task.

An important observation emerges from a study by Hamidi et al.~\cite{hamidi2018interactive}, wherein activation maps are recognized as a valuable tool for analyzing DNNs. Nonetheless, they fall short in providing a comprehensive understanding of how these concepts combine to form final decisions, a limitation shared with LRP. 
Arias-Duart et al.~\cite{arias2022focus} evaluated the robustness of XAI approaches such as LRP, Grad-CAM, and LIME in explaining an image classification task by introducing noise within the image distribution. Their findings showcased Grad-CAM's superior performance across experiments, even when model is not particularly well fit to the task, while LRP performs very well for high performing models, but its reliability diminished with less accurate models. This also seems to be the case of LIME, which suffers from an even larger variance. Researchers came up with new approaches that were extension of CAM to make improvement. For example, Chattopadhay et al.~\cite{chattopadhay2018grad} proposed Grad-CAM++ based on Grad-CAM, this approach like Grad-CAM calculates a weighted combination of positive gradients in the last layer of CNN. However, unlike Grad-CAM, this model calculates gradient weights of pixels instead of features. The authors reported Grad-CAMP++ reached better results in the term of  terms of robustness, human trust, and object localization. 


\textbf{Output summary:} CAM illuminates the pivotal pixels influencing CNN decisions, proving valuable for system verification, debugging, and bias mitigation. Grad-CAM++, however, offers enhanced interpretability and robustness compared to CAM. While all the mentioned methods furnish insights beneficial to AI experts, they may pose challenges for users lacking an AI background as they become hard to be interpreted. Nonetheless, their utilization can foster trustworthiness within the system. It is imperative to note that while CAM provides insights into critical pixels, it fails to elucidate pixel connectivity and correlation, constraining its effectiveness in intervention and output alteration.

\subsection{Further Discussion: Which XAI Method Should We Use to Meet Our Goals }

Our final mapping function that assign XAI methods to their potential users, and properties is presented in Table~\ref{methods}. Our main suggested user for each explanation is highlighted in bold font.

\begin{sidewaystable}[]
\centering
\caption{The proposed mapping function table to map each approach in XAI to users and their desired properties.}
\begin{tabular}{|p{0.13\textwidth}|p{0.2\textwidth}|p{0.25\textwidth}|p{0.2\textwidth}|p{0.07\textwidth}|p{0.1\textwidth}|}
\hline
  
\textbf{XAI Approach} & \textbf{Potential Users} & \textbf{Potential Goals} & \textbf{Desired Properties} & \textbf{Local or Global Explanations} &\textbf{Primary Data Format} \\

\hline
		 CFE~\cite{wachter2017counterfactual} & 1- Stakeholders, 2- Expert Users, 3- \textbf{Non-expert Users}  & 1- System Intervention, 2- Moral Values, 3-  Trustworthiness, 4- Law Requirement  & 1- Plausibility, 2- Model-agnostic, 3- Sparsity, 4- Causality, 5- Run Time, 6- Data Manifold Closeness & Local & 1- Tabular, 2- Image, 3- Text\\

       \hline

    LIME~\cite{ribeiro2016should} & 1- Stakeholders, 2- Expert Users, 3- \textbf{Non-expert Users} & 1- System Intervention, 2- System Debugging, 3- Eliminate Bias, 4- Moral Values, 5-  Trustworthiness, 6- Law Requirement & 1- Model-agnostic, 2- Causality, 3- Run Time & Both  & 1- Tabular, 2- Image, 3- Text\\

		\hline
	
	     SHAP~\cite{lundberg2017unified}  & 1- Stakeholders, \textbf{2- Expert Users}, 3- Non-expert Users & 1- System Intervention, 2- System Debugging, 3- Eliminate Bias, 4- Moral Values, 5-  Trustworthiness, 6- Law Requirement  & 1- Model-agnostic, 2- Causality, 3- Run Time & Both & 1- Tabular, 2- Image, 3- Text\\
  
	\hline

       Saliency Map~\cite{simonyan2013deep} & 1- Stakeholders, \textbf{2- Expert Users}, 3- Non-expert Users & 1- System Intervention, 2- System Debugging, 3- Eliminate Bias, 4- Moral Values, 5-  Trustworthiness, 6- Law Requirement & 1- Model-agnostic, 2- Causality, 3- Run Time & Local& 1- Tabular, 2- Image, 3- Text \\

\hline
		PDP~\cite{friedman2001greedy} & 1- Stakeholders, \textbf{2- Expert Users}, 3- Non-expert Users & 1- System Intervention, 2- System Debugging, 3- Eliminate Bias, 4- Moral Values, 5-  Trustworthiness, 6- Law Requirement & 1- Model-agnostic, 2- Causality, 3- Run Time, 4- Data Manifold Closeness & Global & 1- Tabular \\

  \hline

    ICE	\cite{goldstein2015peeking} & 1- Stakeholders, \textbf{2- Expert Users}, 3- Non-expert Users & 1- System Intervention, 2- System Debugging, 3- Eliminate Bias, 4- Moral Values, 5-  Trustworthiness, 6- Law Requirement & 1- Model-agnostic, 2- Causality, 3- Run Time, 4- Data Manifold Closeness  & Local & 1- Tabular\\

\hline

		ALE~\cite{apley2020visualizing} & 1- Stakeholders, \textbf{2- Expert Users}, 3- Non-expert Users & 1- System Intervention, 2- System Debugging, 3- Eliminate Bias, 4- Moral Values, 5-  Trustworthiness, 6- Law Requirement & 1- Model-agnostic, 2- Causality, 3- Run Time, 4- Data Manifold Closeness & Global & 1- Tabular \\

\hline
		  		
	    LRP~\cite{bach2015pixel} & 1- Stakeholders, \textbf{2- Expert Users}, 3- Non-expert Users & 1- System Intervention, 2- System Debugging, 3- Eliminate Bias, 4- Moral Values, 5-  Trustworthiness, 6- Law Requirement & 1- Model-agnostic, 2- Causality, 3- Run Time & Global& 1- Tabular, 2- Image, 3- Text\\

\hline

        CAM~\cite{zhou2016learning} & 1- Stakeholders, \textbf{2- Expert Users}, 3- Non-expert Users & 1- System Intervention, 2- System Debugging, 3- Eliminate Bias, 4- Moral Values, 5-  Trustworthiness, 6- Law Requirement & 1- Model-agnostic, 2- Causality, 3- Run Time & Local & 1- Tabular, 2- Image, 3- Text\\

\hline

    Grad-CAM~\cite{selvaraju2017grad} & 1- Stakeholders, \textbf{2- Expert Users}, 3- Non-expert Users & 1- System Intervention, 2- System Debugging, 3- Eliminate Bias, 4- Moral Values, 5-  Trustworthiness, 6- Law Requirement & 1- Model-agnostic, 2- Causality, 3- Run Time & Local & 1- Tabular, 2- Image, 3- Text\\

\hline		
\end{tabular}
\label{methods}
\end{sidewaystable}

Explanations can be presented in three formats: visual, text, and numerical, each carrying its own set of advantages and drawbacks. Visual representations excel in quickly conveying a large volume of information, rendering them more easily comprehensible compared to textual or numerical data. However, numerical data tends to offer greater precision and accuracy compared to other formats.
The choice of data format utilized in AI algorithms is another pivotal factor influencing the approach to explanations. Techniques such as LRP and CAM have been primarily designed for image data. In contrast, counterfactual explanations exhibit utility across both tabular and image data, albeit with limited application in text data~\cite{yang2020generating}. Feature importance and saliency maps predominantly serve image data but also find some use in text analysis. PDP, ALE, and ICE methods are commonly deployed for tabular data. LIME and SHAP methods showcase versatility, being adaptable across tabular, image, and text domains.
While LRP and CAM were initially devised to elucidate neural networks operating on image data, there have been constrained endeavors to extend their application to text and tabular datasets. 

In general, the capacity of an XAI algorithm to address causality and correlation between features is pivotal for empowering users to influence the output. However, our study revealed that many algorithms do not adequately account for causality or have not sufficiently developed it. This raises concerns about our ability to fully comprehend AI systems when we lack insights into how one feature impacts others.
Drawing from the summarized information in Table~\ref{methods}, although most XAI approaches can be utilized by non-expert users, they are not tailored specifically for them. Many XAI methods either lack robustness and are vulnerable, requiring close scrutiny from AI experts, or studies have indicated that non-expert users would find them challenging to comprehend.

Another aspect of this discussion pertains to the distinction between the ``ability" of a method to perform a task and the``superiority" of a method in accomplishing that task. For instance, PDP, ALE, and LIME can all be utilized for system debugging, as outlined in Table~\ref{methods}. While each of these methods possesses the ``ability" for debugging, determining the ``superiority" among them requires consideration of factors such as the application, data type, AI algorithm, and user characteristics. For instance, in the context of an object image classification task using a random forest, non-expert users may find it easier to understand a saliency map compared to ALE.

The output of this section is methods that will work better in practice. XAI system developers can use the provided information to select the match method based on their purpose of using XAI and develop properties that are aligned with their intentions.

\section{Future Direction and Open Challenges in XAI} 
\label{short}

So far, we acknowledge that due to the black-box nature of AI algorithms, we are unable to clearly discern why the algorithm reached a specific output. In response to this challenge, researchers have developed explainable AI  to shed light on how algorithms operate.
This section will delve into shortcomings and concerns within the field, proposing potential solutions to address them. Concerns are outlined below:

\subsection{Who Wants XAI?}

The overarching trend in recent papers is the introduction of novel approaches to generating explanations. However, a critical concern emerges regarding the lack of clarity about the intended user of each explanation. It remains uncertain whether the explanation is tailored for an expert, stakeholder, or a non-expert user, such as a patient. The literature lacks well-defined user categories and characteristics for explainable AI, making it challenging to address the distinctive needs of each group. The importance of this concern lies in the understanding that explanations will inherently differ based on the user, necessitating tailored approaches for each category~\cite{brennen2020people}. The specific details of \textbf{how} explanations should vary for each group and \textbf{what} modifications are required for distinct user categories remain unclear~\cite{langer2021we}. This gap in understanding poses a significant challenge in the development and application of XAI methods.


So far, it is evident that various user categories have distinct needs and expectations from XAI, necessitating the establishment of a mapping function to clarify this intricate relationship. 
In certain scenarios, stakeholders often turn CFEs as a means to foster trust among their customers. We can draw a parallel between stakeholders and CFEs, viewing the utilization of CFEs as a method to enhance trustworthiness within the context. Experts and developers would use XAI to debug systems; however, non-expert users, lacking the necessary scientific qualifications, are unlikely to utilize CFEs for debugging purposes~\cite{gilpin2022explanation}. Instead, their primary aim may be to seek explanations that can influence system outputs or to bolster their trust in the system itself. In this scenario, we can align non-expert users with the need for comprehensible explanations and trustworthiness, distinct from the debugging requirements typically associated with CFEs.
Further research can explore refining and expanding the proposed mapping function to determine how these explanations and XAI methods should evolve to be compatible with their users. The study should aim to provide personalized explanations tailored to the distinct needs and goals of each user group.

\subsection{Desired Properties in XAI}

An issue with CFEs and XAI is the varying desirability of different properties in different contexts. For instance, the preference for sparsity implies an inclination towards shorter sentences over longer ones~\cite{verma2020counterfactual}. However, in specific contexts, such as a medical application where users can be doctors and experts, sparsity may not be a desired property, or in some contexts due to technical issue it would be infeasible~\cite{van2022explainable}. In this scenario, doctors may prefer longer sentences that offer more comprehensive information for informed decision-making.

Addressing this concern requires further research to determine the optimal level of sparsity for particular applications and how it can effectively inform users in these contexts. This nuanced approach acknowledges the importance of tailoring explanation properties to the specific needs and preferences of users in diverse domains, ensuring that explanations align with the users' expectations and enhance their decision-making processes.

\subsection{``Good" and ``Bad" Explanations in XAI}

The third missing piece in this puzzle revolves around the observation that some experiments showing that current explanations have no meaningful impact on human decision-making and trust~\cite{alufaisan2020does},~\cite{kenny2021explaining}. However, the crucial question remains unanswered: \textbf{Why} do XAI methods not perform well in practice? \textbf{What changes} are necessary to enhance their effectiveness? Does the lack of meaningful impact apply to all types of users and XAI methods?
It is conceivable that some explanations are more informative for experts than non-experts, suggesting that the meaningful impact may vary across user categories. A case study could help validate or refute this hypothesis.

Another vital issue is the potential for XAI to provide misleading information~\cite{lin2021you},~\cite{mittelstadt2019explaining}. Determining what constitutes misleading information in XAI and how to classify and recognize it are crucial considerations. For instance, could a counterfactual explanation lead a user to take the wrong action? If so, how can misleading information be categorized, such as directly misleading (e.g., recommending a change in the wrong direction or to the wrong feature) or indirectly misleading (e.g., causing the user to draw incorrect conclusions about other features in their profile)?
For example, consider a job applicant who receives two explanations for improving their chances of success in the future. The first advises them to increase their expected salary, which is a direct mistake, while the second suggests increasing job experience and obtaining more recommendations from past employers. The latter could be considered indirectly misleading, as the user may incorrectly conclude that all other aspects of their profile are satisfactory, and they only need to focus on improving their job experience, while changing these specific features is only one way out of many ways to change the output. Addressing these concerns requires careful exploration and research to understand the limitations and potential pitfalls of XAI methods in real-world applications.

\subsection{ Actionability of Features in XAI}

The final challenge in the domain of XAI involves defining immutable features in the field of CFEs. Existing research has introduced algorithms aimed at ensuring that CFEs do not suggest to alter unchangeable features (immutable features), such as race~\cite{verma2020counterfactual},~\cite{karimi2021survey},~\cite{wang2021skyline}. However, the process of determining or defining immutable features remains unclear. For instance, in a loan application scenario, a feature like race may be easily identifiable as immutable. However, in other real-world applications, such as medicine, the list of features may be unknown so determining mutability of a feature would be challenging.

One potential solution to address this challenge is the utilization of time-series data monitoring. By observing data and features over a sufficient period, it becomes possible to identify which features remain unchanged and classify them as immutable, satisfying the actionability criterion. Additionally, this approach allows for the identification of features that change infrequently, meeting the requirement of closeness to the data distribution.
Implementing time-series data monitoring offers a dynamic and adaptive method for defining immutable features in various applications. This approach leverages real-time observations to establish a more accurate and context-specific understanding of which features should be treated as immutable in XAI systems.

\section{Conclusion}	



This paper explores previous works in the field of XAI and XML from a new perspective and categorizes them into three main areas: philosophy, theory, and practice, as discussed in Section~\ref{pre}. We examined how philosophical approaches have developed the vocabulary of explanations and counterfactual explanations within computer science. Theoretical approaches have sought to mathematically capture and formalize explanations. Furthermore, practical perspectives have identified desired properties based on user needs and studied how these explanations can be effectively utilized.

In Section \ref{map}, we delved into XAI users and their motivations. Various methods and their desired properties were then evaluated. The final conclusion of this section introduced a mapping function capable of associating each user, with their different goals, to suitable properties and XAI approaches. This proposed mapping function aims to enable a more personalized and tailored approach in XAI settings.

Finally, in Section~\ref{short}, several shortcomings were identified as follows:
\begin{enumerate}
\item The absence of well-defined user categories for XAI, resulting in a lack of tailored explanations within the field.
\item Current works do not sufficiently account for individual preferences regarding desired properties and explanations.
\item An incomplete understanding of the reasons behind the poor performance of XAI in practical applications.
\item Uncertainty about how to identify immutable and mutable features when there is no prior information about the feature set.
\end{enumerate}

Based on our discussion in this paper, we can conclude that most XAI methods have been developed without careful consideration of their users and the necessary properties these methods should have. Each proposed method should be modified to meet users' needs. The most overlooked users are non-experts who lack knowledge in both AI and the specific domain, and who also have limited flexibility. The proposed mapping function in this work aims to provide a guideline for developers and XAI users in general, to align the characteristics of XAI methods with their intentions and potential users.

Each of the identified shortcomings represents a potential direction for future research. Further investigation can refine and expand the proposed mapping function to determine how explanations and XAI methods should evolve to become more compatible with their users. Such studies should aim to provide personalized explanations tailored to the distinct needs and goals of each user group.


\section*{Declarations}

\textbf{Conflict of interest:} The authors have no conflicts of interest to declare.

\bibliography{sn-bibliography}


\end{document}